%% file: ref.tex
\documentclass[preprint,12pt,authoryear]{elsarticle}

\usepackage{graphicx}
\usepackage{amsmath}
\usepackage{multirow}
\usepackage{pifont}
\usepackage{graphicx}
\usepackage{amsmath}
\usepackage{amssymb}
\usepackage{array}
\usepackage{multirow, nicefrac}
\usepackage[caption=false,font=normalsize,labelfont=sf,textfont=sf]{subfig}

\usepackage{color, xcolor}
\usepackage{multirow}
\usepackage{makecell}
\usepackage{multicol}
\usepackage{bbm}
\usepackage{xcolor}
\usepackage{xspace}
\usepackage{float}
\usepackage{color}
\usepackage{colortbl}
\usepackage[misc]{ifsym}
\usepackage{dsfont}
\usepackage[margin=4pt,font=small,labelfont=bf,labelsep=endash,tableposition=top]{caption}
\usepackage{hyperref}

\newcolumntype{I}{!{\vrule width 3pt}}

\renewcommand{\texttt}[1]{ $ {{\tt #1} } $}  
\usepackage{dsfont}

\newcolumntype{I}{!{\vrule width 3pt}}
\newlength\savedwidth
\newcommand\whline{\noalign{\global\savedwidth\arrayrulewidth
                           \global\arrayrulewidth 2pt}%
                  \hline
                  \noalign{\global\arrayrulewidth\savedwidth}}
\newlength\savewidth
\newcommand\shline{\noalign{\global\savewidth\arrayrulewidth
                           \global\arrayrulewidth 0.5pt}%
                  \hline
                  \noalign{\global\arrayrulewidth\savewidth}}

\usepackage[margin=4pt,font=small,labelfont=bf,labelsep=endash,tableposition=top]{caption}

\def\eg{\emph{e.g., }}

\def\ie{\emph{i.e., }}

\definecolor{mygray}{gray}{.92}

\definecolor{demphcolor}{RGB}{144,144,144}

\definecolor{green}{rgb}{1,0,0}

\newcolumntype{I}{!{\vrule width 3pt}}

\definecolor{mygray}{gray}{.92}

\def\eg{\emph{e.g., }}

\def\ie{\emph{i.e., }}

\usepackage{listings,xcolor} 
\usepackage{tcolorbox}

\journal{Pattern Recognition}

\begin{document}
\include{math_commands}

\begin{frontmatter}



\title{DSText V2: A Comprehensive Video Text Spotting Dataset for Dense and Small Text}


\author{Weijia Wu$^a$, Yiming Zhang$^a$, Yefei He$^a$, Luoming Zhang$^a$, Zhenyu Lou$^a$, Hong Zhou$^a$({\color{blue}{\Letter}}), and Xiang Bai$^b$}

              %
              
\affiliation{organization={Zhejiang University},
            city={HangZhou},
            country={China}}
\affiliation{organization={Huazhong University of Science and Technology},
            city={WuHan},
            country={China}}
            
\begin{abstract}

Recently, video text detection, tracking, and recognition in natural scenes are becoming very popular in the computer vision community. 
However, most existing algorithms and benchmarks focus on common text cases~(\eg normal size, density) and single scenario, while ignoring extreme video text challenges, \ie{} dense and small text in various scenarios.
In this paper, we establish a video text reading benchmark, named DSText V2, which focuses on \textbf{D}ense and \textbf{S}mall text reading challenges in the video with various scenarios.
Compared with the previous datasets, the proposed dataset mainly include three new challenges: 
1) Dense video texts, a new challenge for video text spotters to track and read.
2) High-proportioned small texts, coupled with the blurriness and distortion in the video, will bring further challenges.
3) Various new scenarios, \eg{} `Game', `Sports', etc. 
The proposed DSText V2 includes 140 video clips from 7 open scenarios, supporting three tasks, \ie{} video text detection (Task 1), video text tracking (Task 2), and end-to-end video text spotting (Task 3).
%
%

In this article, we describe detailed statistical information of the dataset, tasks, evaluation protocols, and the results summaries.
Most importantly, a thorough investigation and analysis targeting three unique challenges derived from our dataset are provided, aiming to provide new insights.
Moreover, we hope the benchmark will promise video text research in the community.
DSText v2 is built upon DSText v1, which was previously introduced to organize the ICDAR 2023 competition for dense and small video text.
The dataset website can be found at \href{https://rrc.cvc.uab.es/?ch=22&com=introduction}{\color{blue}{$\tt RRC$}} or \href{https://zenodo.org/records/10010840}{\color{blue}{$\tt Zenodo$}}.

\end{abstract}








\end{frontmatter}


\section{Introduction}

%
The field of reading text from static images has witnessed remarkable progress in recent years, thanks to advancements in deep learning and the availability of extensive public datasets such as 
MJSynth~\cite{jaderberg2014synthetic},
SynthText~\cite{synthtext}, ICDAR2015~\cite{karatzas2015icdar}, FlowText~\cite{zhao2023flowtext}, and Total-Text~\cite{totaltext}, as well as various algorithms including PSENet~\cite{psenet}, EAST \cite{zhou2017east}, MaskTextSpotter \cite{lyu2018mask}, 
,
Polygon-Free~\cite{wu2022polygon}
,
DText~\cite{cai2022arbitrarily}, TextMountain~\cite{zhu2021textmountain}, TextCohesion~\cite{wu2019textcohesion}, and FOTS~\cite{liu2018fots}.
In contrast, the progress in video-level text spotting~\cite{yin2016text} has been notably slow, which limited numerous applications of video text, \eg{video understanding~\cite{srivastava2015unsupervised}, video retrieval~\cite{dong2021dual,wu2023large}, video text translation, and license plate recognition~\cite{anagnostopoulos2008license}, etc.}
There have been a few previous video text spotting benchmarks attempting to develop video text spotting, which focuses on easy cases, \eg{} normal text size, and density in a single scenario.
ICDAR2015~(Text in Videos)~\cite{karatzas2015icdar}, as the most popular benchmark, was introduced during the ICDAR Robust Reading Competition in 2015 focus on wild scenarios: walking outdoors, searching for a shop in a shopping street, etc. 
YouTube Video Text~(YVT)~\cite{nguyen2014video} contains 30 videos from YouTube. The text category mainly includes overlay text~(caption) and scene text (\eg{driving signs, business signs}).
RoadText-1K~\cite{reddy2020roadtext} provides 1,000 driving videos, which promote driver assistance and self-driving systems.
LSVTD~\cite{cheng2019you} proposes 100 text videos, 13 indoor (\eg{ bookstore, shopping mall}) and 9 outdoor (\eg{highway, city road}) scenarios, and support two languages, \ie{} English and Chinese.
BOVText~\cite{wu2021bilingual} establishes a large-scale, bilingual video text benchmark, including abundant text types, \ie{} title, caption, or scene text.

\begin{figure*}[t]
\begin{center}
\includegraphics[width=0.98\textwidth]{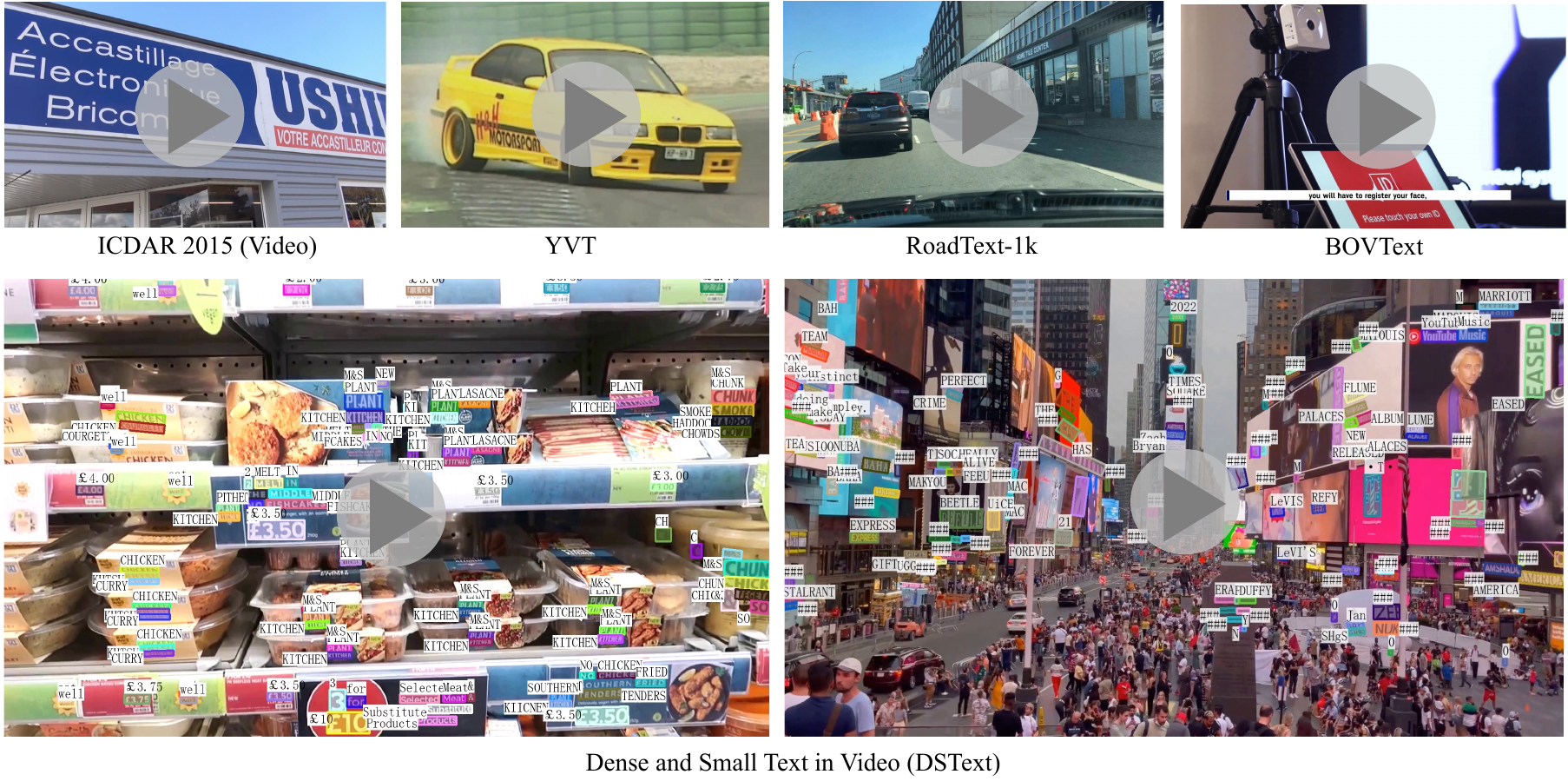}
\caption{\textbf{Visualization of DSText.} Different from previous benchmarks, DSText focuses on dense and small text challenges. Each frame of the video contains dense small texts, coupled with camera movements, posing unique challenges for detection, tracking, and recognition models.}
\label{fig1_vis}
\end{center}
\end{figure*}

However, as shown in Figure~\ref{fig1_vis}, the above benchmarks still suffer from some limitations: 1) Most text instances present standard text size without challenge, \eg{} ICDAR2015(video) YVT, BOVText.
2) Sparse text density in a single scenario, \eg{} RoadText-1k, and YVT, which can not evaluate the small and dense text robustness of the algorithm effectively.
3) Most benchmarks present unsatisfactory maintenance. YVT, RoadText-1k, and BOVText all do not launch a corresponding competition and release open-source evaluation scripts.
Besides, the download links of YVT even have become invalid.
The poor maintenance plan is not helpful to the development of video text tasks in the community.
To break these limitations, in this work, we establish one new benchmark, which focuses on dense and small texts in various scenarios, as shown in Figure~\ref{fig1_vis}.
Our dataset has several advantages and unique challenges.
\textbf{Firstly},
the high-quality, high-resolution, and various videos with dense and small texts~(\ie{140 videos, 62.1k video frames, and 2.2m text instances}) are collected from \textit{YouTube} enabling the development of deep design specific for video text spotting.
\textbf{Secondly}, 
Each frame of the video contains a significant amount of small text, which presents significant challenges for detection, tracking, and recognition models when encountering distortions and motion blur.
\textbf{Thirdly}, 
The average text density per frame reaches a high value of $24$, significantly surpassing the maximum text density of previous datasets ($5.55$ from ICDAR 2015). This poses significant challenges for tracking models, as the high text density can lead to ID switches in tracking, resulting in a decrease in tracking accuracy.
The benchmark mainly supports three tasks, \ie{} \textit{video text detection}, \textit{video text tracking}, and end-to-end \textit{video text spotting} tasks, including 140 videos with 62.1k frames.

To further advance the field, we also organize the ICDAR 2023 Video Text Reading competition for dense and small text, which includes two competition tracks: video text tracking, and spotting tasks. 
This competition can serve as a standard benchmark for assessing the robustness of algorithms that are designed for video text spotting in complex natural scenes,
which is more challenging.
%
In this paper, we have expanded the dataset with an additional 40 videos, resulting in a total of 140 videos. 
And we also provide comprehensive data analysis, experimental analysis, and additional insights into the unique challenges posed by our dataset.
In the Section 3 of this paper, we provide a detailed overview of how the V2 version of the dataset was constructed, annotated, and present comprehensive information on the data distribution and statistical analysis. 
In the Section 4, we list all tasks and metrics, along with a brief analysis. 
Section 5 includes various experimental comparisons and introduces new insights.

The main contributions of this work are three folds:
\begin{itemize}
    \item We collect and annotate a high-quality, high-resolution video text benchmark with various video domains, which includes $140$ videos, $62.1k$ video frames, and $2.2m$ text instances.
    
    \item  Compared to the current existing video text reading datasets, the proposed DSText V2 provides some \textit{unique} features and challenges, including 1) Abundant video scenarios, high-quality videos, 2) Massive and high-proportion of small text, and 3) dense text distribution per frame. 
    
    \item  
    We provide comprehensive data analysis, experimental analysis, and additional insights into the unique challenges of our dataset, enabling future researchers to better understand and leverage its potential.
\end{itemize}

\section{Related Work}
In this section, we provide a concise overview of the relevant literature and benchmarks related to our work, specifically focusing on end-to-end text-spotting methods and corresponding benchmarks.
\subsection{Image Text Spotting}
Numerous image-level deep learning-based approaches~\cite{li2017towards,he2018end,lyu2018mask,liu2020abcnet,liu2023towards,lu2021master,zheng2020scale,wu2020synthetic,wu2020texts,wu2023diffumask} have been introduced to tackle the task of image text spots, resulting in significant performance improvements.
\cite{li2017towards} pioneered the development of an end-to-end trainable scene text spotting method. 
Their approach effectively integrates detection and recognition features using RoI Pooling~\cite{ren2015faster}, achieving notable success.
\cite{lyu2018mask} introduced a novel approach called Mask TextSpotter, which extends the capabilities of Mask R-CNN by incorporating character-level supervision for simultaneous character detection and recognition. %
In addition, to mature algorithms, there are also numerous excellent datasets~\cite{zhou2015icdar,veit2016coco,totaltext,karatzas2013icdar} available that contribute significantly to the advancement of research in this field. 
These datasets provide a diverse range of annotated images, offering valuable resources for training and evaluating text detection and recognition models.
Some notable datasets such as ICDAR 2015~\cite{zhou2015icdar} benchmarks and COCOText~\cite{veit2016coco}, which offer a wide variety of text instances in different contexts.
This data is collected using Google Glasses, which captures a wide range of diverse scenes, including but not limited to street views, indoor environments, and shopping malls.
Furthermore, there are emerging datasets like Total-Text~\cite{totaltext} that focus on curved and irregular text, pushing the boundaries of existing algorithms and fostering innovation in the field of text analysis.

\subsection{Video Text Spotting}

Unlike image text detection, the development of video-level tracking in the context of text spotting has been relatively slow. 
In recent years, there have been several efforts~\cite{yu2021end,tu2018hierarchical,yin2016text,rong2014scene,wu2021bilingual,wang2017end} to address the challenges of video text tracking and spotting.
Researchers have explored various approaches to overcome the complexities associated with tracking text in videos, including visual feature association~\cite{yu2021end}, learnable query embedding~\cite{wu2021bilingual}, detection bounding box association~\cite{rong2014scene}.
\cite{nguyen2014video} performs character detection and recognition via scanning-window templates trained with mixture models.
\cite{yu2021end} leverage the advantages of contrastive learning to track text across consecutive frames. They employ a contrastive loss to minimize the distance between the embeddings of the same text instance across frames while maximizing the distance between embeddings of different text instances.
\cite{wang2017end} propose a method to link text across adjacent frames using a combination of detection box IoU and transcription-based edit distance matching.
%
%
\cite{cheng2019you} and Free~\cite{cheng2020free} propose a video text spotting framework that performs text recognition only once for the localized text. This approach reduces the computational costs and improves efficiency.
\cite{wu2021bilingual} propose a video text spotting approach that utilizes object features from the previous frame as a tracking query for the current frame. They employ IoU matching to obtain accurate text tracking and train a separate recognition model to obtain the final recognition results.
TransDETR~\cite{wu2022end} was the first to propose an end-to-end framework that addresses the tasks of detection, tracking, and recognition in a unified framework. This method employs a query embedding to represent a text instance across multiple frames, effectively modeling the long sequence relationship. 
Overall, end-to-end video text spotting remains scarce and there is still significant room for improvement in terms of speed and accuracy.

\begin{table*}[t]
    \centering
	\caption{\textbf{Statistical Comparison.} `Box', and `Text Area' denote the box annotation type and average area  (\# pixels) of text while the shorter side of the image is 720 pixels. `Text Density' refers to the average text number per frame. 
 }
	\label{table1}
	\input{table/table1.tex}

\end{table*}

\subsection{Video Text Dataset}
The progress in video text spotting has been limited in recent years due to the scarcity of efficient datasets. 
The ICDAR 2015 Video dataset~\cite{zhou2015icdar} comprises only 25 training videos and 25 test videos. The videos are categorized into a few specific scenarios, such as walking outdoors or searching for a shop in a street, where the majority of the text instances are considered relatively easy cases, as they do not pose significant challenges in terms of text size, density, or other characteristics.
The Minetto Dataset~\cite{minetto2011snoopertrack} is a relatively small dataset that includes only 5 videos captured in outdoor scenes. The frame size of the videos is 640 x 480 pixels.
YVT~\cite{nguyen2014video}, consists of a total of 30 videos. Out of these, 15 videos are used for training, and the remaining 15 videos are designated for testing. 
RoadText-1K~\cite{reddy2020roadtext} offers a collection of driving videos consisting of 1000 videos, which are sampled from the BDD100K~\cite{yu2018bdd100k}. The dataset primarily focuses on the driving scenario, making it specific to road scenes.
BOVText~\cite{wu2021bilingual}, is a bilingual dataset collected from YouTube and Kuaishou. It comprises over 2,000 videos. A majority of these videos in this dataset may not exhibit significant challenges in terms of text spotting, as shown in Figure~\ref{fig1_vis}.
Existing datasets primarily focus on normal text and lack challenging examples of dense and small text. Besides, many datasets suffer from poor maintenance. 

\begin{figure*}[t]
\begin{center}
\includegraphics[width=0.99\textwidth]{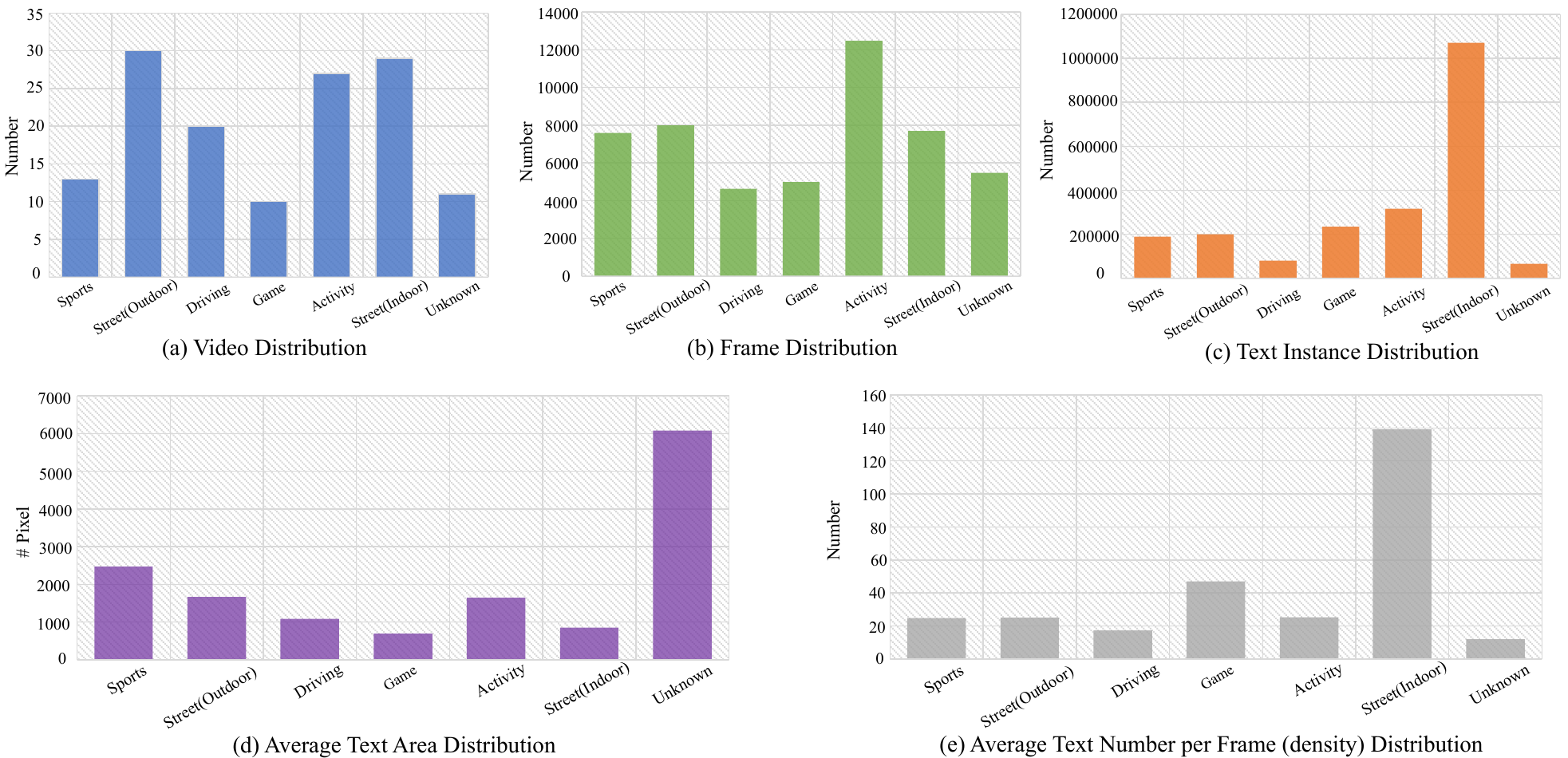}
\caption{\textbf{The Data Distribution for $7$ Open Scenarios.} (a) Video number distribution (b). The number distribution of video frames. (c) Text instance distribution. (d) Average text area~(\# pixels) distribution while the shorter side of the image is $720$. (e) The distribution of average text number per frame~(density).}
\label{fig4_vis}
\end{center}
\end{figure*}

\section{Dataset}

\subsection{Dataset and Annotations}
\subsubsection{Dataset Source.} We collect high-quality videos with dense and small text from three data sources: 
\begin{itemize}
    \item 
    1) $30$ videos sampled from the large-scale video text dataset BOVText~\cite{wu2021bilingual}.
BOVText as the largest video text dataset, comprising over $2,000$ videos, is a valuable data source. 
It covers a wide range of scenarios and includes a substantial number of videos with small and dense text.
We employ a selection process to identify the top $30$ videos with small and dense texts based on criteria that the average text area within the video and the average number of text instances per frame. 
This ensures that we focus on videos that exhibit a high concentration of small and dense text.  

\item 2) We collect $10$ videos for driving scenario from RoadText-1k~\cite{reddy2020roadtext}. 
As shown in Figure~\ref{table1}, the RoadText-1k dataset exhibits a significant presence of small texts, which aligns with the challenge of our dataset.  
Thus we also randomly select $10$ videos to enrich the driving scenario.

\item 3) $100$ videos for other scenarios, such as street view scenes and supermarkets, are collected from YouTube.
Except for BOVText and RoadText-1k, we also need more high-quality videos with dense and small texts for other scenarios, such as games, street view scenes, and supermarkets. Therefore, we also collect $100$ videos with dense and small texts from YouTube.

\end{itemize}
Therefore, we obtain $140$ videos with $62.1$k video frames, as shown in Table~\ref{table1}.
Then the dataset is randomly divided into two parts: the training set with $34,907$ frames from $90$ videos, and the testing set with $27,234$ frames from $50$ videos.
In this paper, we have added an additional $40$ videos to the training set in DSText expanding the size and diversity of the dataset.
Furthermore, the original $12$ video scenarios have been reorganized into $7$ more representative scenarios.
This categorization involved grouping certain categories with fewer data samples~(\ie{} movie, news) into the "unknown" category.

\begin{figure}
\begin{center}
\includegraphics[width=0.65\textwidth]{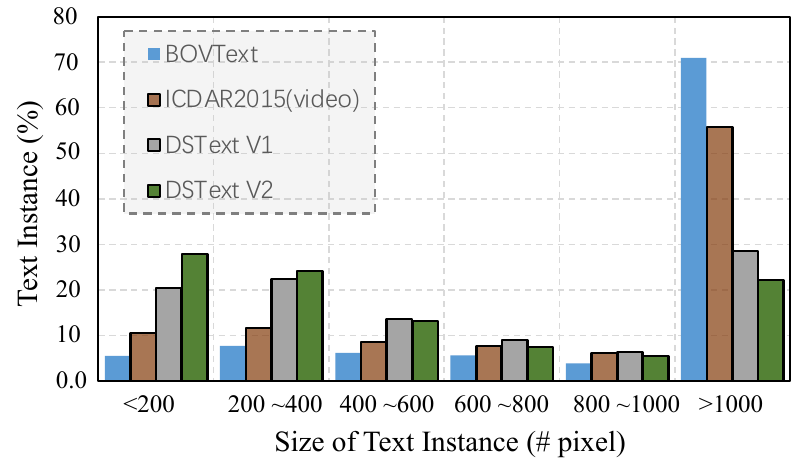}
\caption{\textbf{The distribution of different text size range on different datasets} "\%" denotes the percentage of text size region over the whole data. Text area (\# pixels) is calculated while the shorter side of the image is 720 pixels.}
\label{fig5_vis}
\end{center}
\end{figure}

\begin{figure}[t]
\begin{center}
\includegraphics[width=0.75\textwidth]{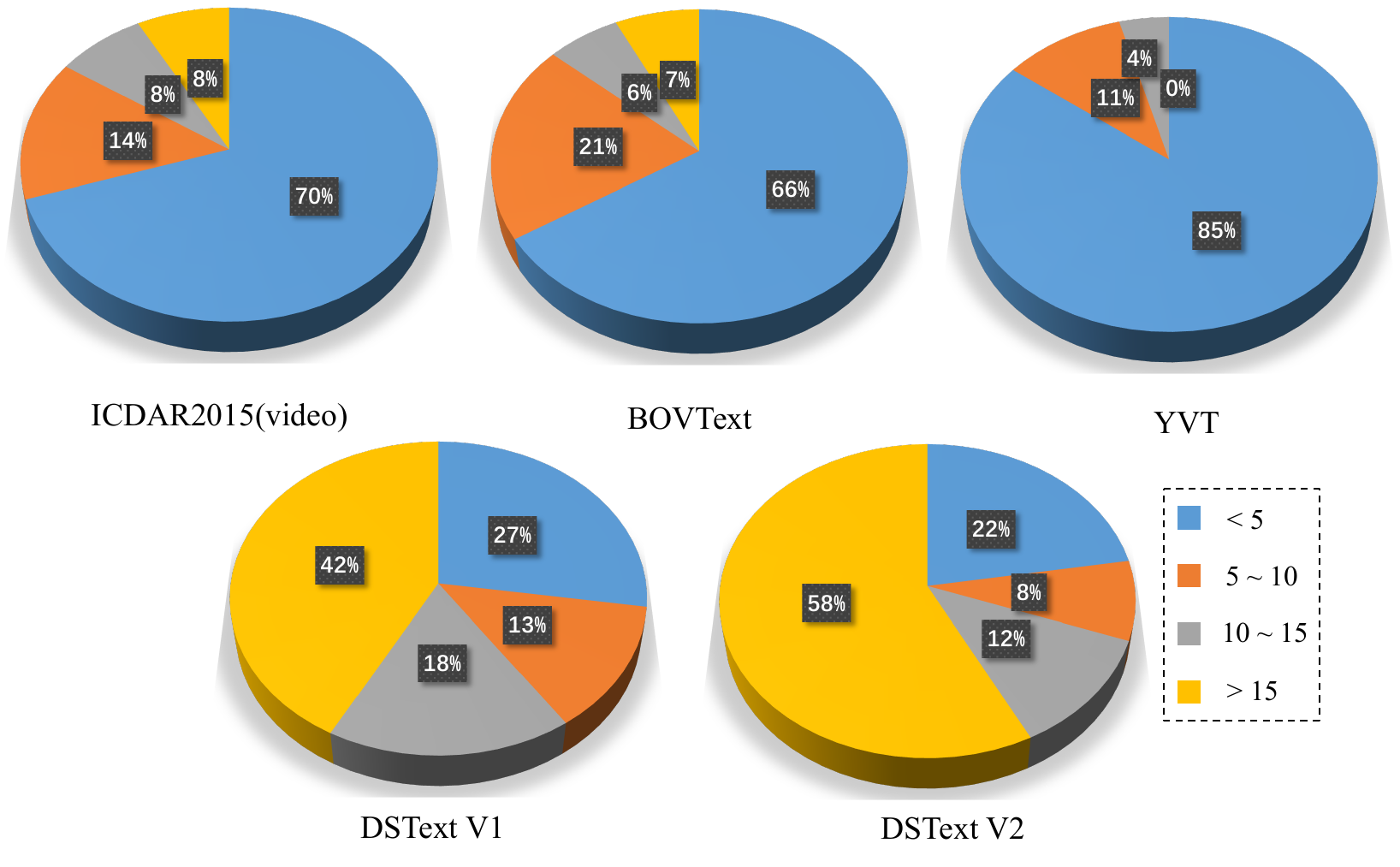}
\vspace{-2mm} 
\caption{\textbf{Comparison for frame number of different text numbers.} "\%" denotes the percentage of the corresponding frame over the whole data. The majority of video frames from the existing datasets only contain 0 to 5 text instances. In contrast, our dataset includes a significant number of text-dense scenes, with $58\%$ of video frames containing 15 or more text instances.}
\label{fig5_vis1}
\end{center}
\vspace{-3mm} 
\end{figure}

\begin{figure}[t]
\begin{center}
\includegraphics[width=0.94\textwidth]{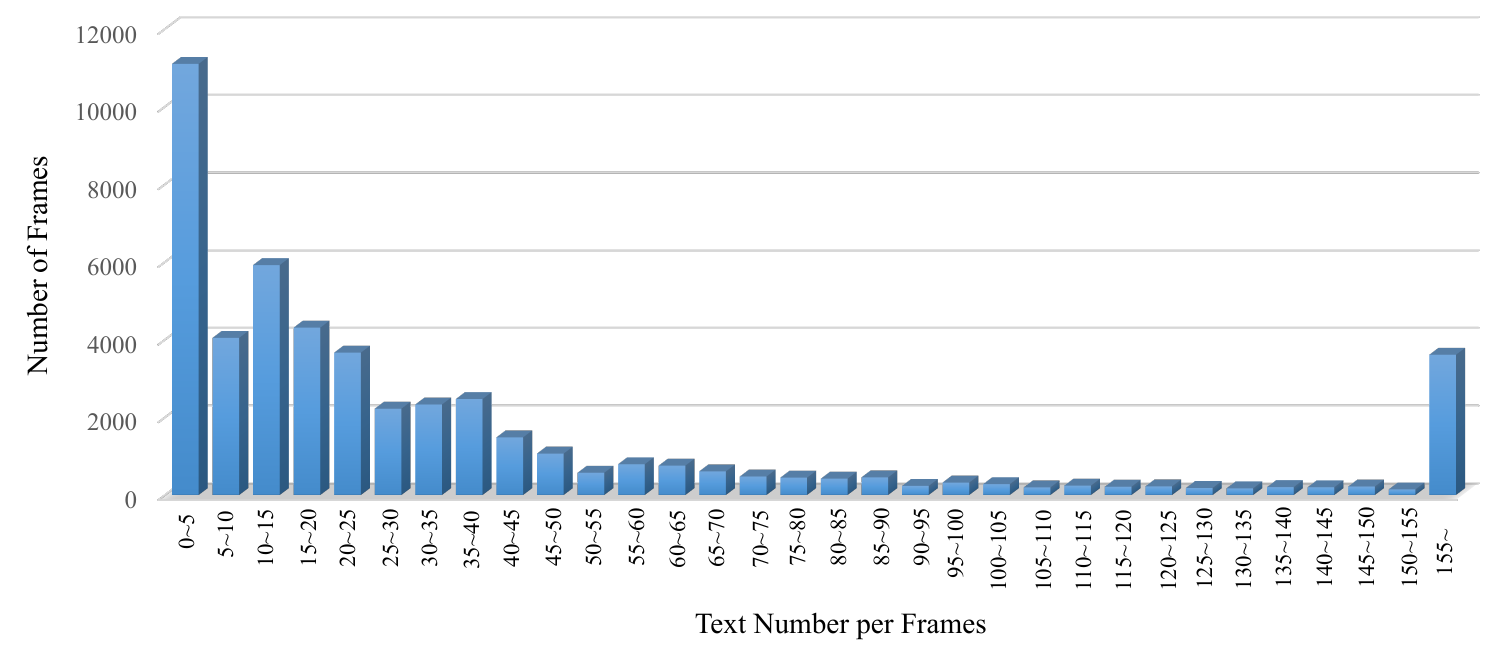}
\caption{\textbf{Distribution of frame for different text numbers on DSText V2.} Frames with more than 10 texts account for around $70\%$ of the dataset.}
\label{fig5_vis11}
\end{center}
\vspace{-3mm} 
\end{figure}

\begin{figure}
\begin{center}
\includegraphics[width=0.96\textwidth]{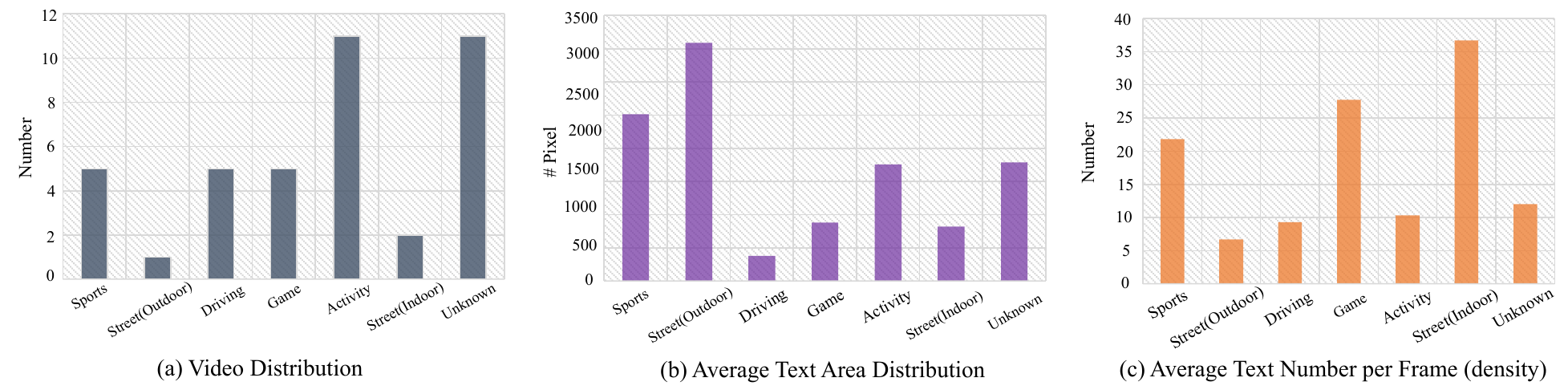}
\caption{\textbf{The Data Distribution for  the newly added 40 videos on DSText V2.}}
\label{add40}
\end{center}
\vspace{-3mm} 
\end{figure}

\begin{figure}
\begin{center}
\includegraphics[width=0.85\textwidth]{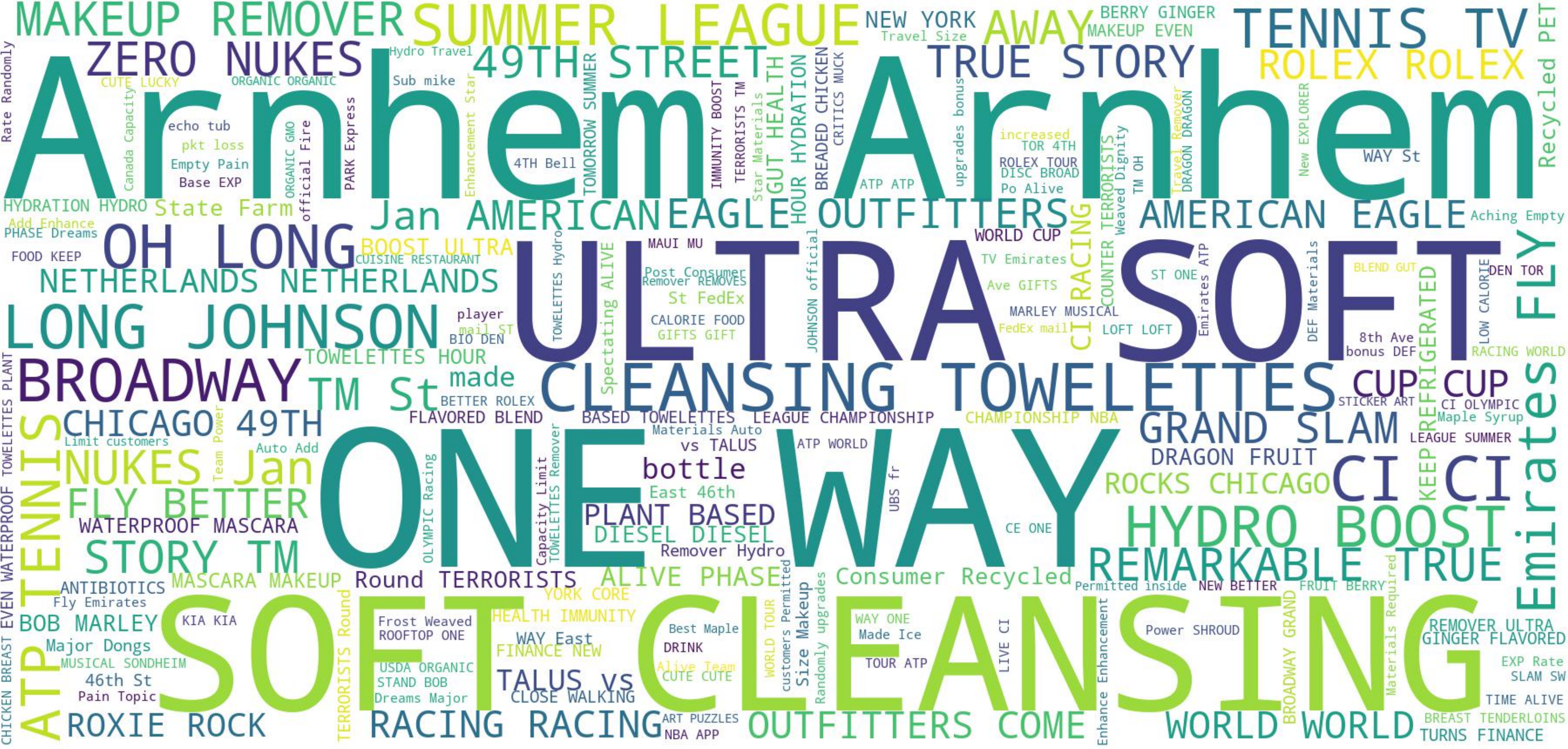}
\caption{\textbf{Wordcloud visualizations for DSText V1.}}
\label{fig5_vis112}
\end{center}
\vspace{-3mm} 
\end{figure}

\subsubsection{Annotation} 
We adopt two main annotation approaches based on the different sources of data:
1) For the $30$ videos from BOVText, we just adopt the original annotation, which includes four kinds of description information: the rotated bounding box of detection, the tracking identification(ID) of the same text, the content of the text for recognition, the category of text, \ie{} caption, title, scene text, or others.
2) As for the $110$ videos from RoadText-1k and YouTube, we hire a professional annotation team to label each text for each frame.
The annotation format is the same as BOVText.
For each video, we extract frames at a rate of 15 frames per second (FPS) and require each annotator to annotate frame by frame.
Then, we invite an audit team with around 5 persons to carry out another round of annotation checks, and re-label part video frames with unqualified annotation.
We require a bounding box and text transcription accuracy of over $95\%$ for acceptance.
This means that all the correctly annotated boxes should cover the entire text region, and any missed or incorrect annotations are considered annotation errors.
As for text transcription, every letter should be accurately transcribed.
Similar to the ICDAR2015 video dataset~\cite{zhou2015icdar}, for blurry or non-English texts, we require the annotators to only annotate the bounding box and tracking ID while setting the text transcription as `ignored'. 
These texts will not be considered when calculating tracking or spotting metrics. 
In other words, the detection of these texts will not receive any reward in terms of metrics, and there will be no penalty for missing them.
One mentionable point is that the videos from RoadText-1k only provide the upright bounding box~(two points), thus we abandon the original annotation and annotate these videos with the oriented bounding box.
%
%
As a \textit{labor-intensive} job, the whole labeling process took \textbf{30} men in one a and a half months, \ie{} around \textbf{7,200} man-hours, to complete the 110 video frame annotations.
In our dataset, due to the high density of text in each frame, the average annotation cost per frame is significantly higher compared to other datasets, requiring approximately three to four times more time and effort.
As shown in Figure~\ref{fig1_vis_case}, it is quite time-consuming and expensive to annotate a mass of text instances at each frame.

\subsection{Dataset Comparison and Analysis}

In this section, we present detailed statistical data and comparative analysis.
Table~\ref{fig4_vis} presents an overall comparison for the basic information, \eg{} the number of videos, frame, text, average text area, text density, and supported scenarios.
In comparison with previous works, the proposed DSText V2 shows the denser text instances density per frame ~(\ie{} average $42.4$ texts per frame) and smaller text size~(\ie{} average $1,758$ pixels area of texts).
Figure~\ref{fig5_vis}, Figure~\ref{fig4_vis}, and Figure~\ref{fig5_vis1} present detailed data distributions and comparisons with the existing datasets for DSText v2.
In addition, we have provided comprehensive data analytics for the newly added 40 videos in DSText V2, as illustrated in Figure~\ref{add40}.
This includes statistics on the distribution of video quantities per video scenario, the average text area distribution per video scenario, and the average text count (per frame) distribution per video scenario category.
Due to the substantial presence of Street (indoor) and Street (outdoor) scenes in the V1 version, in the V2 version, we primarily increased the number of videos in other scene categories, such as Activity and Sports scenes.
It is noteworthy that all videos of Unknown video scenario are newly added; this scenario category was not present in the V1 version.

\subsubsection{Video Scenario Attribute}
As shown in Figure~\ref{fig4_vis}, we present the detailed distribution of video, frame, and frame of $7$ open scenarios and an "Unknown" scenario on DSText V2.
`Street View~(Outdoor)' and `Sport' scenarios present most video and text numbers, respectively.
And the frame number of each scenario is almost the same.
It is worth mentioning that the text in the Street (Indoor) scenario exhibits extremely high text density, with an average of around $140$ texts per frame, as shown in Figure~\ref{fig4_vis} (e).
This poses a significant challenge, particularly for transformer-based architectures with a limited capacity of handling 100 queries, such as TransDETR~\cite{wu2022end} (which can output only 100 bounding boxes). 
Additionally, the text in this scenario is also extremely small, with an average pixel area size of less than 1000, which is much smaller than existing datasets.
We also present more visualizations for `Game', `Driving', `Sports', and `Street View' in Figure~\ref{fig3_vis}.

\subsubsection{Higher Proportion of Small Text}
Figure~\ref{fig5_vis} presents the proportion of different text areas for different datasets. 
The proportion of big text~(more than $1,000$ pixel area) on our DSText V2 is less than that of BOVText and ICDAR2015(video) with at least $20\%$.
Moreover, it is also about $5\%$ smaller than DSText V1, which is due to the inclusion of smaller-sized text in our V2 version.
DSText V2 also presents a higher proportion for small texts~(less 400 pixels) with up to $50\%$.
This type of small text poses significant challenges, as demonstrated in the supermarket (street indoor) scene shown in Figure~\ref{fig3_vis}. 
For text detectors and recognizers, such text is prone to false negatives, false positives, or recognition errors.
RoadText-1K~\cite{reddy2020roadtext} and LSVTD~\cite{cheng2019you} also exhibit relatively small average text areas in Table~\ref{table1}. However, their text density is quite sparse, with only $0.75$ texts and $5.12$ texts per frame, which is significantly lower than our text density of $42.4$.
Besides, RoadText-1k only focuses on the driving domain, which limits the evaluation of other scenarios.

\begin{figure}
\begin{center}
\includegraphics[width=0.85\textwidth]{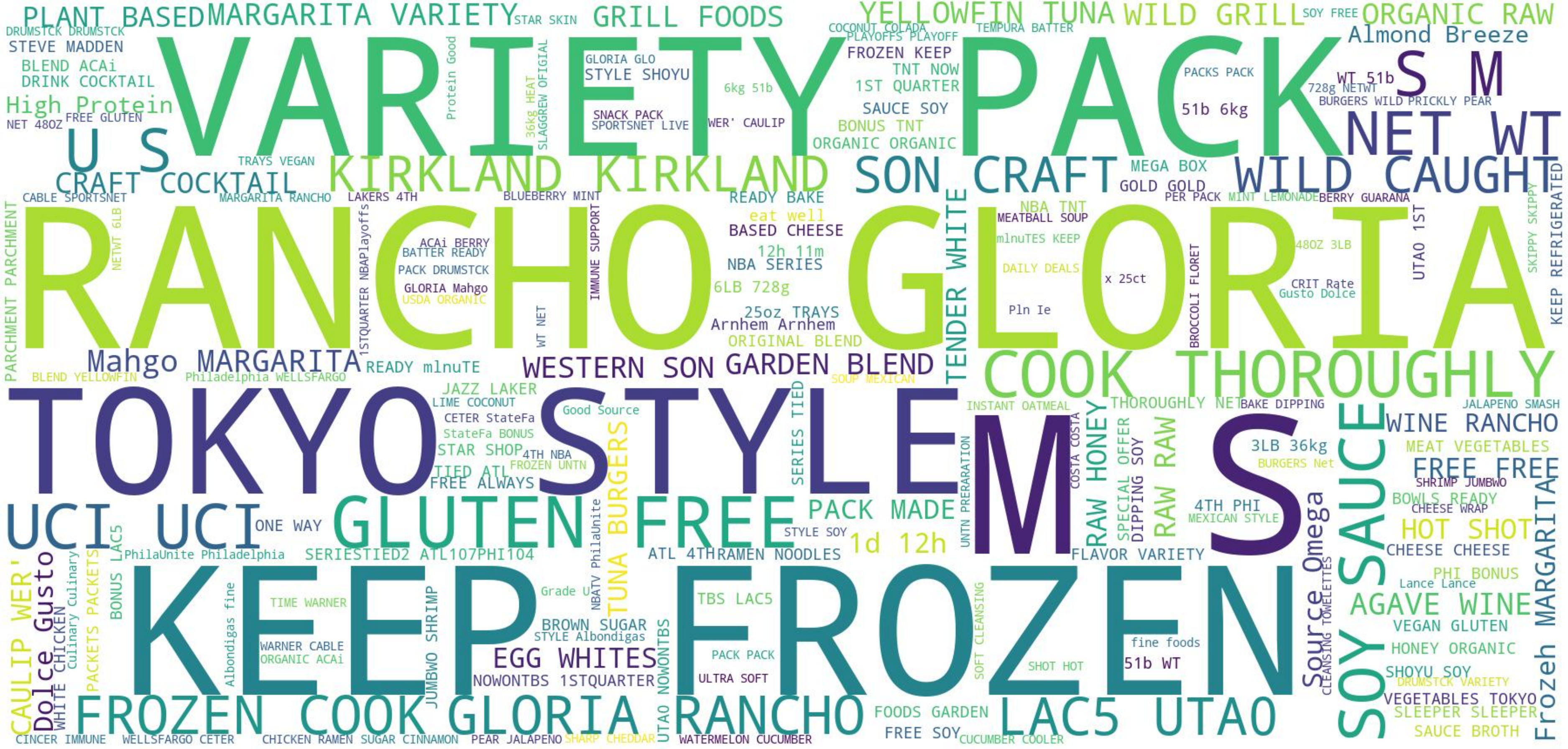}
\caption{\textbf{Wordcloud visualizations for DSText V2.}}
\label{fig5_vis111}
\end{center}
\vspace{-4mm} 
\end{figure}

\begin{figure*}[t]
\begin{center}:
\includegraphics[width=1\textwidth]{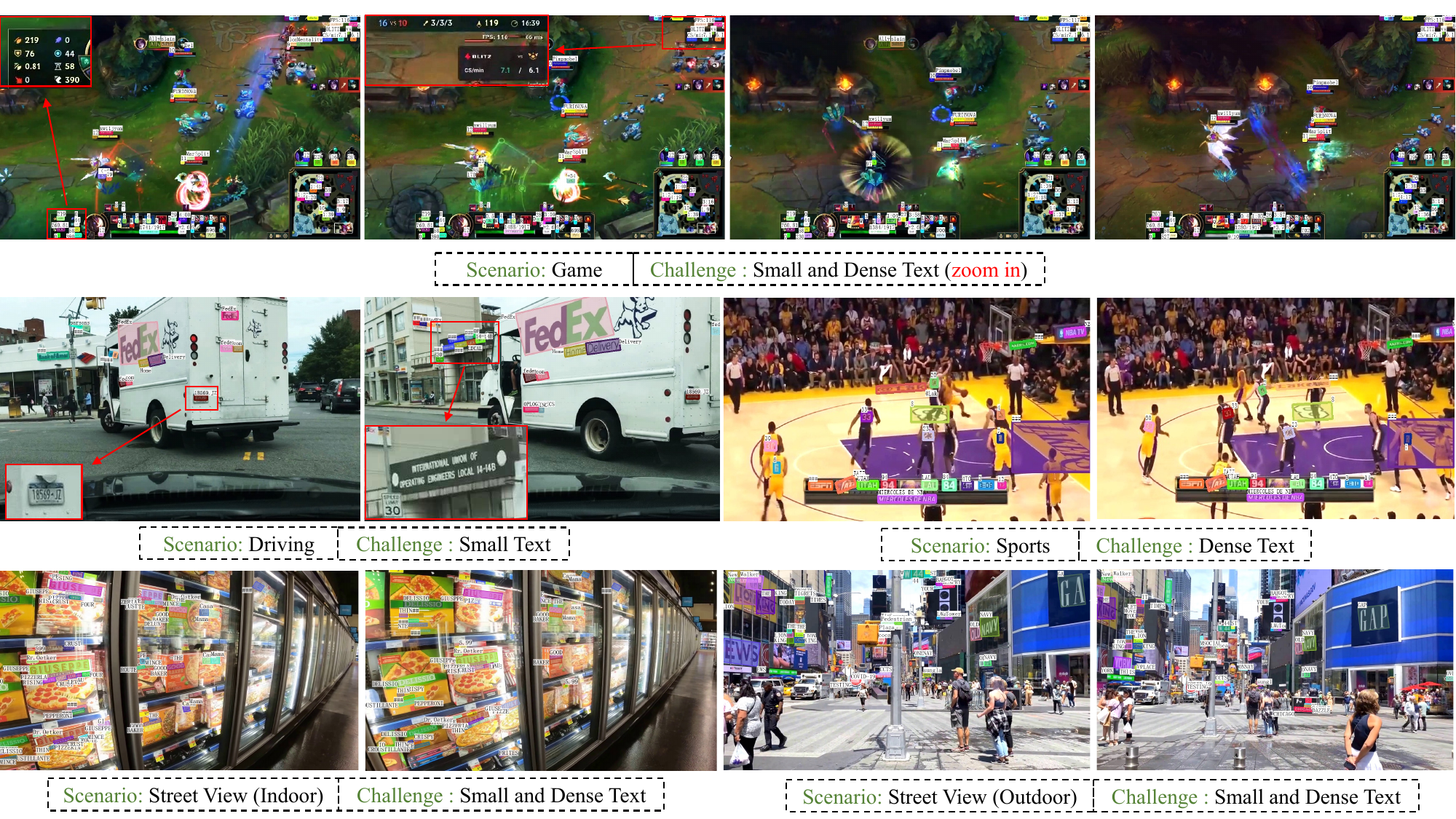}
\vspace{-5mm} 
\caption{\textbf{More Qualitative Video Text Visualization of DSText V2.} DSText V2 covers small and dense texts in various scenarios, which is more challenging.}
\label{fig3_vis}
\end{center}
\vspace{-3mm} 
\end{figure*}

\subsubsection{Dense Text Distribution}
\begin{figure*}[t]
\begin{center}:
\includegraphics[width=0.98\textwidth]{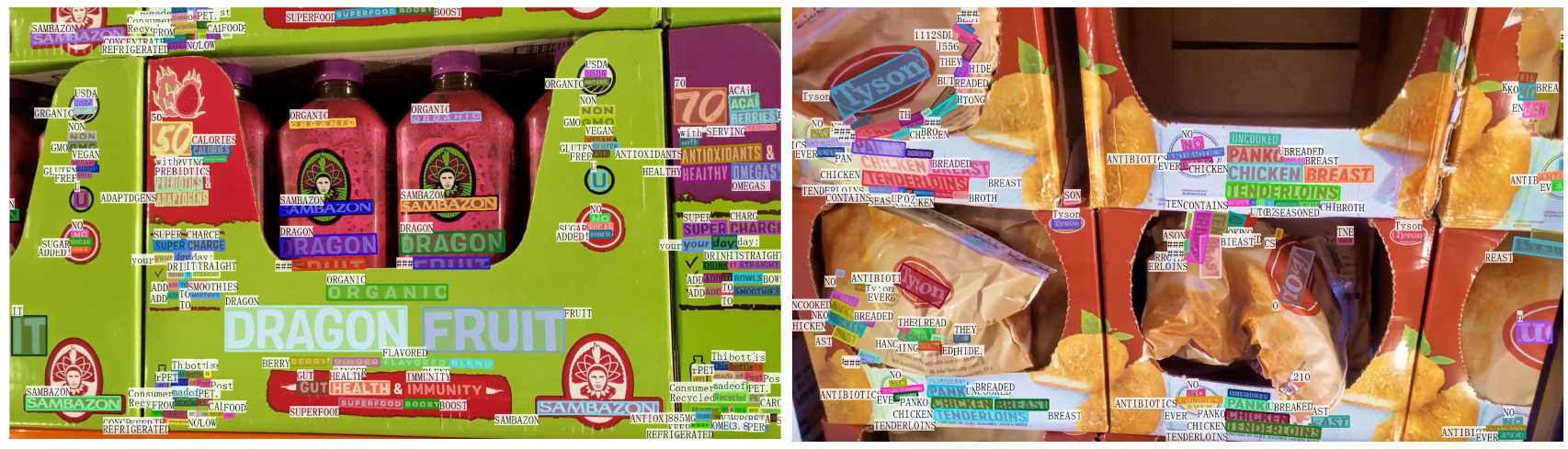}
\vspace{-2mm} 
\caption{\textbf{One Case for around 200 Texts per Frames.} DSText V2 includes huge amounts of small and dense text scenarios.}
\label{fig1_vis_case}
\end{center}
\vspace{-3mm} 
\end{figure*}

Figure~\ref{fig5_vis1} presents the comparison of text density at each frame.
The frames with more than $15$ text instances occupies $58\%$ in our dataset, at least $30\%$ improvement than the previous work~($7\%$ and $8\%$ for BOVText and ICDAR2015 video), which presents more dense text scenarios.
Besides, the proportion of frames with fewer than $5$ text instances is significantly lower in our dataset, accounting for only $22\%$.
In contrast, ICDAR2015 video had $70\%$ and BOVText had $66\%$ of frames with fewer than $5$ text instances.
%
%
Therefore, the proposed DSText V2 shows the challenge of dense text tracking and recognition. 
Figure~\ref{fig5_vis11} shows the detailed data distribution for our DSText V2.
Frames containing 5-40 text instances comprise the majority, while there are also over 4,000 frames with densely packed scenes containing more than 150 text instances.
More visualization can be found in Figure~\ref{fig3_vis}~(Visualization for various scenarios) and Figure~\ref{fig1_vis_case} (Representative case with around 200 texts per frame).

\subsubsection{WordCloud}
We also visualize the word cloud for text content in Figure~\ref{fig5_vis111} and Figure~\ref{fig5_vis112}.
All words from annotation must contain at least $3$ characters, we consider the words less than four characters usually insignificant, \eg{} `is'.
Comparing Figure~\ref{fig5_vis112} and Figure~\ref{fig5_vis111}, we can observe significant changes in the word cloud distribution due to the addition of the extra 40 videos. 
Some new words have emerged, such as "variety" and "PACK."
%


\section{Tasks and Metric}
The proposed dataset supports three tasks: 1) video text detection, where the objective is to obtain a rough estimation of the text areas in the image, in terms of bounding boxes that correspond to parts of text. 2) video text tracking, where the objective is to localize and track all words in the video sequences. and 3) the end-to-end video text spotting: where the objective is to localize, track and recognize all words in the video sequence.
In this paper, we will conduct a comprehensive evaluation of these three tasks.

\label{metric}
\subsection{Task 1: Video Text Detection}
Video text detection is similar to image-level detection, the method requires detecting all the text in the video frames and returning the corresponding detection boxes.
And we will directly use the image-level evaluation metrics from ICDAR 2015~\cite{karatzas2015icdar} to evaluate.
\subsection{Task 2: Video Text Tracking}
%
%
%
The task requires one network to detect and track text over the video sequence simultaneously.
Given an input video, the network should produce two results: a rotated detection box, and tracking ids of the same text.
For simplicity, we adopt the evaluation method from the ICDAR 2015 Video Robust Reading competition~\cite{zhou2015icdar} for the task.
The evaluation was revised in 2020, utilizing MOTChallenge~\cite{dendorfer2019cvpr19} for multiple object tracking
%
For each method, MOTChallenge provides three different metrics: the Multiple Object Tracking Precision (MOTP), the Multiple Object Tracking Accuracy (MOTA), and the IDF1. See the 2013 competition report~\cite{karatzas2013icdar} and MOTChallenge~\cite{dendorfer2019cvpr19} for details about these metrics.
%
%
%
%

\begin{table}[t]
    \centering
	\caption{\textbf{The Related Metric for Video Text Spotting Task}. These metrics have been adopted by ICDAR2015(V), ICDAR2013(V), RoadText-1k, BOVText, DSText V1, and DSText V2.}
\label{metric_table}
\vspace{-2mm}
 \input{table/metric}
 \vspace{-3mm}
\end{table}

\subsection{Task 3: End-to-End Video Text Spotting}
Video Text Spotting~(VTS) task that requires simultaneously detecting, tracking, and recognizing text in the video. 
The word recognition performance is evaluated by simply whether a word recognition result is completely correct. And the word recognition evaluation is case-insensitive and accent-insensitive. All non-alphanumeric characters are not taken into account, including decimal points, such as `1.9' will be transferred to `19‘ in our GT. 
Similarly, the evaluation method~(\ie{} ${\rm ID_{F1}}$, ${\rm MOTA}$ and ${\rm MOTP}$) from the ICDAR 2015 Robust Reading competition is also adopted for the task.
Table~\ref{metric_table} shows all metrics and their corresponding explanations.
The main issue with MOTA is its primary focus on the number of incorrect decisions made by the tracker, such as ID switches (IDSW). 
In certain scenarios, there is a greater concern for the duration of tracking for a specific ID.
For instance, when a trajectory is consistently tracked for the majority of its lifespan, even if it experiences multiple ID switches in the early or late stages, we consider the trajectory to be a reasonable prediction. 
Therefore, ${\rm ID_{F1}}$ is introduced to address this phenomenon.
${\rm ID_{F1}}$ is particularly concerned with the length of time a tracker correctly matches the same ID.
In cases where a trajectory maintains correct ID matches for the majority of its tracking duration, occasional acceptable ID switches may occur.
Conversely, even if there is only one ID switch, but the majority of ID matches are incorrect, this is deemed unacceptable in practical applications.
%


\section{Experimental}
\subsection{Implementation Details for Baseline}


\begin{figure*}
\begin{center}
\includegraphics[width=0.9\textwidth]{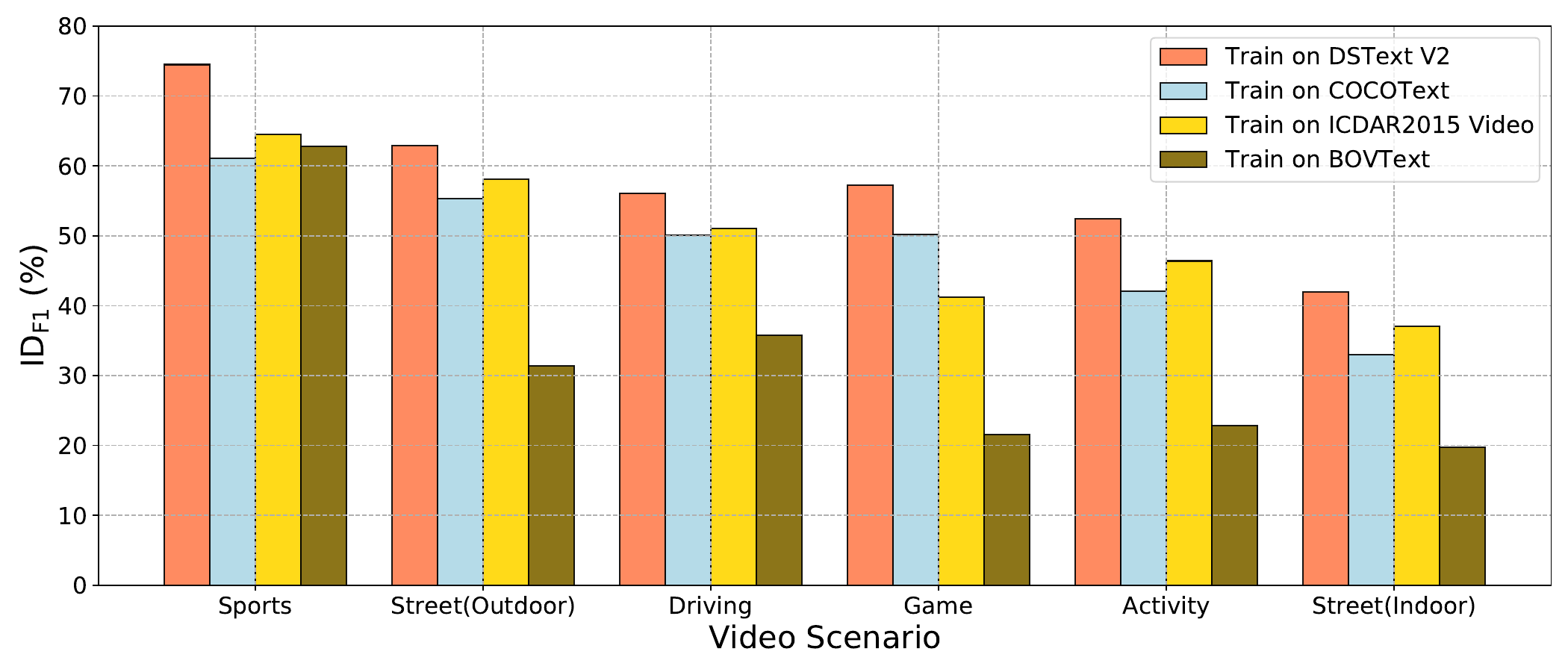}
\vspace{-2mm}
\caption{\textbf{Tracking performance (\ie{} ${\rm ID_{F1}}$) with TransDETR~\cite{wu2022end} in different scenarios of DSText V2.} The "Street Indoor" scenario presents smaller text areas and denser text data, which introduces significant challenges for existing algorithms and datasets.}
\label{ablation_cros}
\end{center}
\vspace{-3mm} 
\end{figure*}

%
\textbf{TransDETR~\cite{wu2022end}}, as the official baseline~\footnote{https://rrc.cvc.uab.es/?ch=22\&com=downloads} of DSText V1, is used to further evaluate and validate our dataset.
TransDETR is a novel, simple, and end-to-end video text DEtection, Tracking, and Recognition framework~(TransDETR), which views the video text spotting task as a direct long-sequence temporal modeling problem.
We followed the official open-source code~\footnote{https://github.com/weijiawu/TransDETR} for various implementation details, including learning rate, data augmentation, and optimizer.  All speed and performance are tested with a
batch size of 1 on a V100 GPU and a 2.20GHz CPU in a single thread.

\textbf{Other baselines.} 
In addition to the TransDETR, we also included several other baselines. As shown in Table~\ref{spotting_table}, image-level detectors, including EAST~\cite{zhou2017east}, and PSENet~\cite{psenet}, were used to detect text objects.
We strictly followed the settings and configurations provided in the official open-source code for all aspects of object detection. 
\cite{wang2017end} was utilized to link and match text objects across frames using IOU and edit distance. 
Finally, CRNN~\cite{shi2016end} was employed as the recognition model to identify each text trajectory in the video.
We used the re-implementation code of DTRB~\footnote{https://github.com/clovaai/deep-text-recognition-benchmark} for our experiments.

\begin{figure}
\begin{center}
\includegraphics[width=0.8\textwidth]{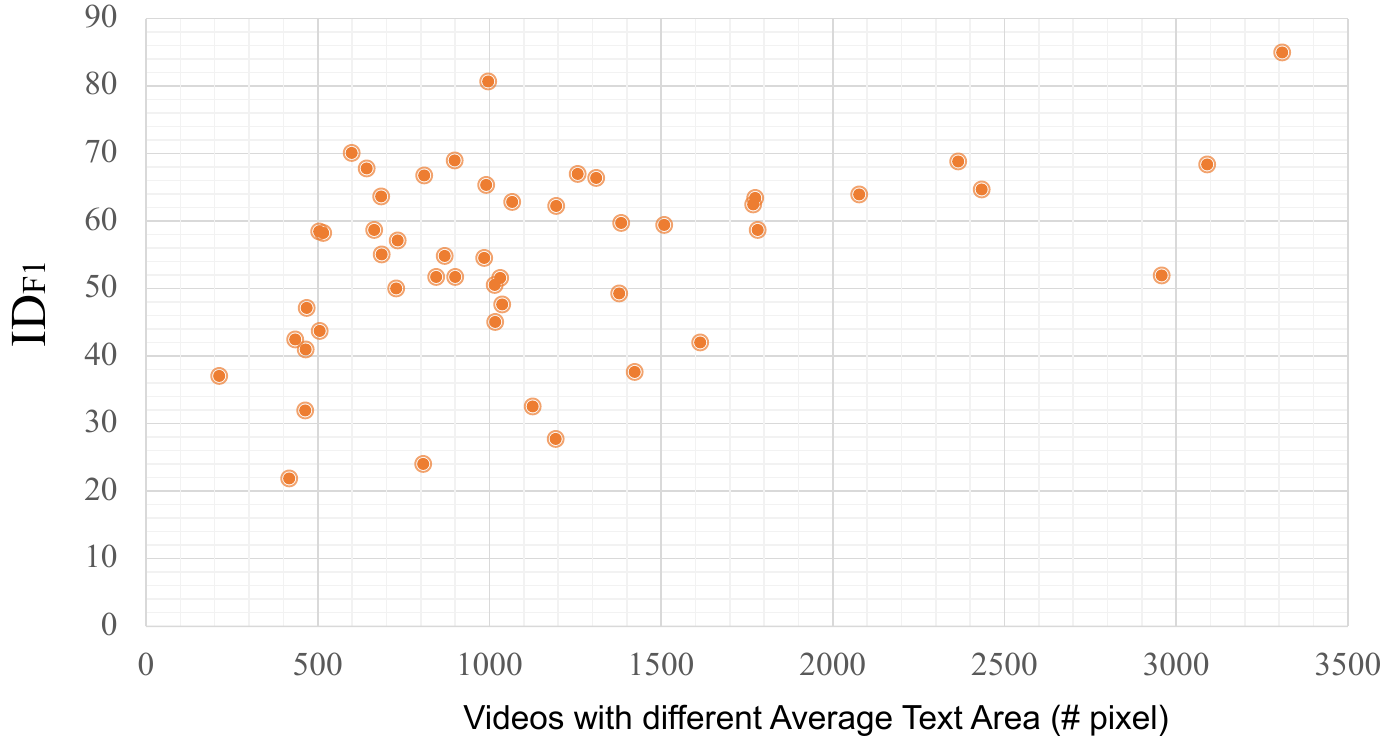}
\caption{\textbf{Tracking performance (\ie{} ${\rm ID_{F1}}$) of videos with different average text area.} TransDETR is the baseline, and is trained on the training set of DSText V2.}
\label{text_size}
\end{center}
\end{figure}

\subsection{Attribute Experiments Analysis for DSText V2}
\textbf{Small Text, New Challenge for Video Text}. 
A key contribution of this work is the distribution of a high proportion of small text. To validate the technical challenges posed by small text, we analyze the problem from two perspectives. Firstly, we compare the performance of videos with different text densities.
As shown in Figure~\ref{text_size},
we compute the average text area and corresponding ${\rm ID_{F1}}$ metric for 50 test videos.
It can be observed that within a certain range, larger text areas result in better tracking performance.
However, when the average text area is less than 500 pixels, it becomes challenging to achieve an ${\rm ID_{F1}}$ score above $50\%$.
Secondly, we evaluate the tracking performance based on video scenarios.
As shown in Figure~\ref{ablation_cros},
we categorize the evaluation based on video scenarios.
It is evident that scenarios with smaller average text areas exhibit lower tracking performance. For example, the "Street Indoor" scenario presents the lowest performance, with an average text area of only 900 pixels. 
Additionally, we explore the transferability of models trained on other existing video text benchmarks and tested on our dataset. 
Figure~\ref{ablation_cros} presents that the challenge of small text persists, even when models are trained on other benchmarks.

\begin{figure}
\begin{center}
\includegraphics[width=0.8\textwidth]{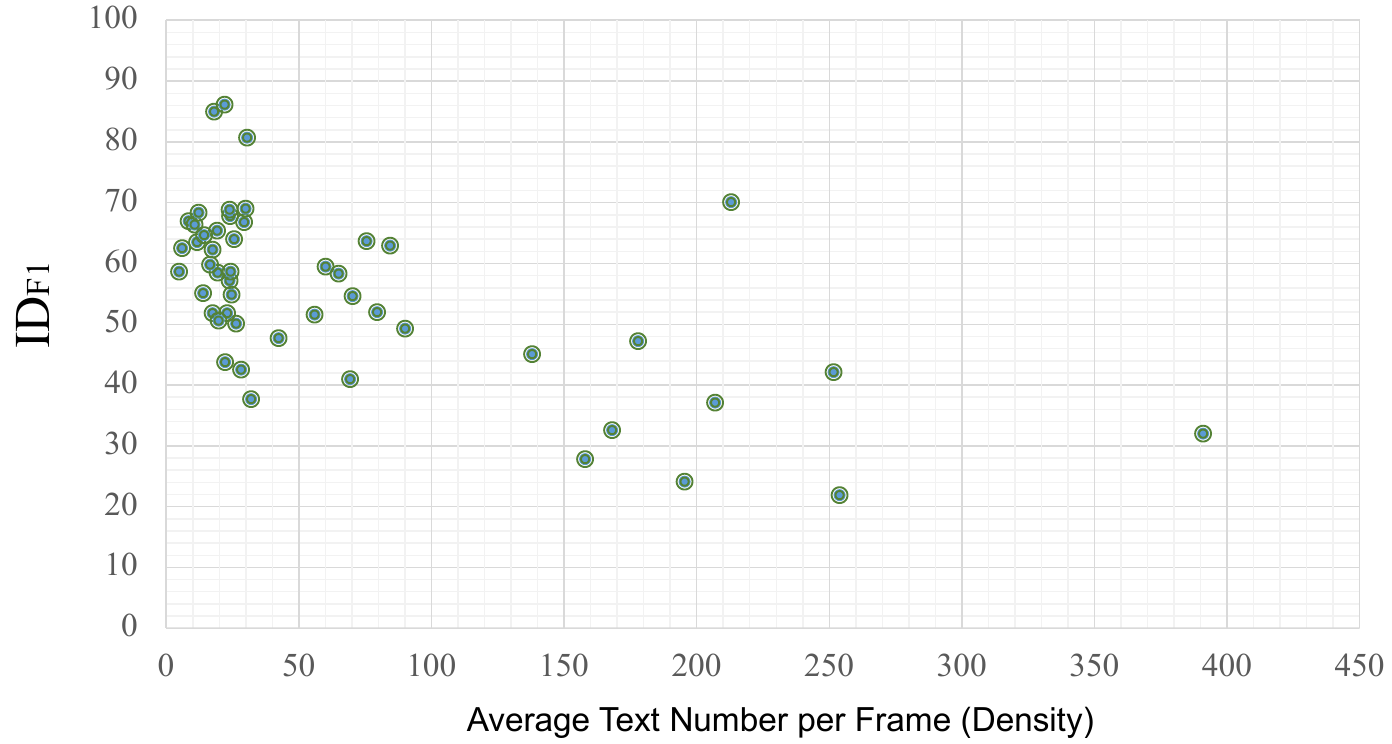}
\caption{\textbf{Tracking performance (\ie{} ${\rm ID_{F1}}$) of videos with different text density.} TransDETR is the baseline, and is trained on the training set of DSText V2.}
\label{text_number}
\end{center}
\end{figure}

\begin{figure}
\begin{center}
\includegraphics[width=0.7\textwidth]{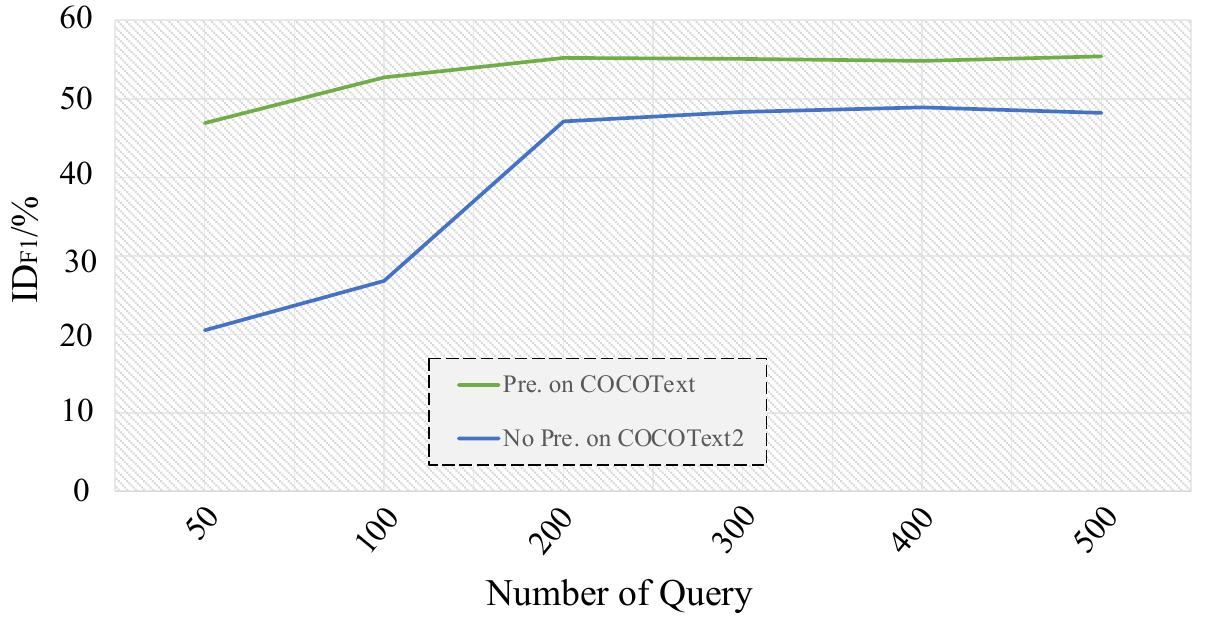}
\vspace{-2mm} 
\caption{\textbf{Ablation Study for the Number of Query.} When the number of queries is less than 200, increasing the query quantity results in significant performance gains.}
\label{query}
\end{center}
\vspace{-3mm} 
\end{figure}

\textbf{Dense Texts, New Challenge for Video Text}.
Similar to analyzing the challenge posed by small text, we also examined the distributional changes between text density and ${\rm ID_{F1}}$ metric.
As shown in Figure~\ref{text_number}, it is evident that as the text density increases, the ${\rm ID_{F1}}$ performance gradually decreases. 
Interestingly, when the average text density is less than 100, almost no video performance is below $30\%$. 
However, among videos with an average text density greater than 150, three of them exhibit performance below $30\%$.

\textbf{Effect of Different Scenarios}
Figure~\ref{ablation_cros} compares the performance on the test set across different scenarios.
The “Sport" scenario demonstrates the best performance, while the “Street Indoor" scenario performs the worst. 
This is reasonable since the “Street Indoor" scenario contains a large amount of dense small text, with an average text area of only $900$ pixels.
However, the average text per frame is as high as $140$.
It is important to note that the “Unknown" scenario does not exist in the test set. The “Unknown" scenario is introduced as a new scenario in the training set of the V2 version, and we did not expand it to the test set.

\textbf{Effect of the Number of Query.}
Table~\ref{query} presents the ablation study for the number of query.
When the number of queries is less than 200, increasing the query quantity results in significant performance gains. 
However, when the number of queries exceeds 200, further increases seem to yield diminishing returns, indicating a saturation in performance.

\subsection{Video Text Detection}
As shown in Table~\ref{detection_table}, we provide six baselines for evaluation. 
EAST, PSENet, and DB are image-level detectors, while TransDETR is an end-to-end video text spotter.
For TransDETR, we directly use the tracking detection boxes as predictions to calculate the metrics. 
With 200 queries, TransDETR achieves the highest performance with an F-measure of $62.9\%$.
In comparison to existing benchmarks, the detection performance is noticeably lower.
For example, the highest performance on ICDAR15 is close to $90\%$, while the current highest performance on the video version of ICDAR15 is $75\%$.
There still seems to be significant room for improvement.
Furthermore, it is worth noting that dense text poses a significant challenge not only in terms of detection accuracy but also in terms of detection speed.
For algorithms that cannot process text instances in parallel, such as DB and PSENet, the inference speed is greatly restricted due to the need to process over 100 text instances in each frame.
As a result, the inference speed is only around 5 frames per second (fps), which is unacceptable for video tasks.

\begin{table*}[t]
    \centering
	\caption{\textbf{Video Text Detection Performance on DSText V2}. `Size' and `Q' means the shorter side of the input and query, respectively.}
\label{detection_table}
 \input{table/detection.tex}
\end{table*}

\begin{table*}[t]
    \centering
	\caption{\textbf{Video Text Tracking Performance on DSText V2}. `Size' and `Q' means the shorter side of the input and learnable query embedding, respectively.  Mostly Tracked (${\rm M\mbox{-}T}$) denotes the number of tracked objects at least 80\% of lifespan, Mostly
    Lost (${\rm M\mbox{-}L}$) denotes the number of objects tracked less than 20\% of lifespan. EAST~\cite{zhou2017east}, VMFT~\cite{wang2017end}, TransDETR~\cite{wu2022end}, and PSENet~\cite{psenet} are used as the baselines.}
\label{track_table}
 \input{table/tracking.tex}
\end{table*}

\begin{table*}[t]
    \centering
	\caption{\textbf{End-to-End Video Text Spotting Performance on DSText V2}. `Q' means the number of query.  EAST~\cite{zhou2017east}, VMFT~\cite{wang2017end}, TransDETR~\cite{wu2022end}, CRNN~\cite{shi2016end}, and PSENet~\cite{psenet} are used as the baselines.}
\label{spotting_table}
 \input{table/spotting}
\end{table*}

\begin{table*}[t]
    \centering
	\caption{\textbf{Cross-dataset Evaluations on DSText V2}. TransDETR with 200 queries is used as the baseline.}
\label{generalizablity_table}
 \input{table/generalizablity.tex}
\end{table*}

\begin{table*}[t]
    \centering
	\caption{\textbf{ Detailed information on the size, density, and tracking performance of text in each video.} Dense and small text scenarios are typically more challenging.}
\label{video_test_size}
 \input{table/video_test_size}
\end{table*}

\subsection{Video Text Tracking}
Table~\ref{track_table} presents the corresponding tracking performance for DSText V2.
With 200 queries, TransDETR achieves the best performance with an IDF1 of $55.2\%$ and a MOTA of $39.3\%$, showing a significant improvement compared to a simple combination of image text detectors and tracking link algorithms.
It's worth noting that compared to YORO and TransVTSpotter, our method has a simpler and more efficient pipeline, eliminating the need for multiple models to assist each other.
Additionally, pretraining on COCOText~\cite{veit2016coco} leads to a significant improvement, with at least a $10\%$ increase in ${\rm ID_{F1}}$.
Furthermore, increasing the number of queries results in further improvement.
This is reasonable since, as shown in Figure~\ref{fig5_vis11}, there are approximately $4000$ frames with over $150$ texts.
In such scenarios, having more queries is necessary to capture all the texts accurately.
We also provide more detailed information for each video  Figure~\ref{video_test_size}.

\subsection{Video Text Spotting}

We also provide the performance of end-to-end video text spotting in Table~\ref{spotting_table}. 
Compared to the text tracking task, the performance of the end-to-end task appears to be significantly unsatisfactory.
One main reason is that, unlike tracking metrics, the end-to-end task requires both tracking and recognizing the text with complete accuracy.
The predicted result is considered correct only when every letter in each word is correctly recognized.
However, this is a challenging task, especially in video scenarios where there are camera movements, distortions, and occlusions.
Overall, with 200 queries and pretraining on COCOText~\cite{veit2016coco}, the model achieves the best performance with an ${\rm ID_{F1}}$ of $25.5\%$ and a MOTA of $-1.3\%$. 
This indicates that there is still significant room for improvement, such as using more powerful recognition models, among other approaches.
Additionally, further iterations are needed to enhance the inference speed in dense scenes.
The current inference speed is insufficient for practical applications, especially for video streaming applications.

\subsection{Cross-datasets Evaluations}
Table~\ref{generalizablity_table} presents the related cross-datasets evaluations for studying the  generalizability and transferability between different domains.
The ICDAR2015 video dataset, being the most popular dataset currently, is primarily used as a benchmark for comparison.
It is observed that a model trained on the ICDAR2015 video dataset fails to achieve satisfactory performance on DSText V2, yielding only a 49.6 ${\rm ID_{F1}}$. 
However, a model trained on DSText V2 achieves a notable 63.4 ${\rm ID_{F1}}$ score on the ICDAR2015 video dataset.
This suggests that the model trained on DSText V2 possesses stronger generalization capabilities, performing well on other domain data.
The dissatisfactory performance of the model trained on ICDAR2015 video in DSText V2 indicates the presence of unique challenges in the video scenarios of DSText V2 that are absent in the ICDAR2015 video dataset.
This is reasonable, as DSText V2 presents numerous challenges related to small and densely packed text.
In contrast, the majority of text in the ICDAR 2015 video dataset corresponds to typical scenes, as indicated by the data comparison in Table~\ref{table1}.
Therefore, a model trained on ICDAR 2015 video may struggle to handle very small and dense text scenarios.

\section{Potential Limitations}
Although our dataset addresses important gaps and challenges, there are still some potential limitations.
Here, we briefly enumerate and discuss several limitations.
Firstly, the dataset mainly focuses on multi-oriented English text and does not support curved text annotation.
While multi-oriented English text is the most common occurrence in real-life scenarios, the inability to support curved text is a limitation.
This is particularly relevant in scenarios where curved text, such as text on billboards, cannot be adequately fit by multi-oriented bounding boxes.
Furthermore, the scale of this dataset is still not very large. 
Exploring ways to further expand the dataset size at a lower cost remains one of the directions worth considering.

\section{Conclusion and Future Work}
Here, we present a new video text reading benchmark, which focuses on dense and small video text.
We provide a well-maintained project page~\footnote{https://rrc.cvc.uab.es/?ch=22\&com=introduction} with corresponding links for dataset download.
Compare with the previous datasets, the proposed dataset mainly includes two new challenges for dense and small video text spotting. 
High-proportioned small texts are a new challenge for the existing video text methods.
Furthermore, we provide a comprehensive data analysis that includes the number of videos in different scenarios, the average text area, and the average text density. 
Additionally, we provide more experimental analysis corresponding to the two unique challenges, namely small text and dense text, further validating their impact.
%
%
%
Overall, we believe and hope the benchmark, as one standard benchmark, develops and improve the video text tasks in the community.



\section{Acknowledgements} This work is supported by the National Key Research and Development Program of China (\# 2022YFC3602601),  the Key Research and Development Program of Zhejiang Province of China (\# 2021C02037),
and the National Natural Science Foundation (NSFC\#62225603)
Besides, we extend our appreciation to OpenAI for the valuable contribution of ChatGPT, which played a crucial role in refining the grammar and enhancing the writing of this paper.

\bibliographystyle{elsarticle-num-names}
\bibliography{ref}

\end{document}

%% file: table/table1.tex
\def\x{{$\footnotesize \times$}}
\scriptsize  
\setlength{\tabcolsep}{1.0pt}
\begin{tabular}{l|c|c|c|c|p{0.07\columnwidth}|p{0.08\columnwidth}|p{0.24\columnwidth}}
    \whline
    
	Dataset &  Video & Frame & Text & Box & Text Area & Text Density & Supported Scenario~(Domain) \cr\shline \hline
	\hline
    AcTiV-D~\cite{DBLP:conf/das/ZayeneSTHIA16}& 8 & 1,843 & 5,133 & Upright& - & - & News video
    \cr\cline{8-8} 
    YVT~\cite{nguyen2014video}  & 30 & 13k & 16k & Upright& 8,664 & 1.15 & Cartoon, Outdoor(supermarket, shopping street, driving...) 
    \cr\cline{8-8} 
    ICD15 VT\cite{zhou2015icdar} & 51 & 27k & 144k & Oriented & 5,013 & 5.33 &  Driving, Supermarket, Shopping street...
    \cr\cline{8-8} 
    RoadText\cite{reddy2020roadtext} & 1k & 300k & 1.2m & Upright & 2,141 & 0.75 & Driving
    \cr\cline{8-8} 
    LSVTD\cite{cheng2019you} & 100 & 66k & 569k & Oriented & 2,254 & 5.52 & Shopping mall, Supermarket, Hotel...
    \cr\cline{8-8} 
    BOVText~\cite{wu2021bilingual} & 2k & 1.7m & 8.8m & Oriented & 10,309 & 5.12 &    Cartoon,~Vlog, Travel, Game, Sport, News ...
    \cr\cline{8-8} 
    DSText V1~\cite{wu2023icdar} & 100 & 56.0k & 671k & Oriented & 1,984 & 23.5 &  Driving, Activity, Street View~(indoor), Street View~(outdoor), Sports, News, Movie, Cooking\\
    \hline
    DSText V2 & 140 & 62.1k & 2.2m & Oriented & \textbf{1,758} & \textbf{42.4} &  Driving, Activity, Street View~(indoor), Street View~(outdoor), Game, Sports, Unknown\\
    \whline
     
\end{tabular}

%% file: table/metric.tex
\def\x{{$\footnotesize \times$}}
\scriptsize
\setlength{\tabcolsep}{2pt}
\begin{tabular}{c|c|p{0.68\columnwidth}}
    \whline
	Metric   & Better &Description   
    \cr\shline \hline
	\hline

   ${\rm ID_{F1}}$ & higher 
   & ${\rm ID_{F1}}$ Score~\cite{ristani2016performance}. The ratio of correctly identified detections\&recognitions over the average number of ground-truth and computed detections\&recognitions. Formula: $\frac{{\rm  ID_{Recall}} \times {\rm  ID_{Precision}} \times 2}{{\rm  ID_{Recall}} + {\rm  ID_{Precision}}}$    \cr \hline
   
   ${\rm  ID_{Recall}}$
   & higher
   & Ratio of correct detections\&recognitions to total number of GTs. Formula: $\frac{{\rm  TP}}{{\rm  TP} + {\rm  FN}}$ \cr \hline
    ${\rm  ID_{Precision}}$
   & higher 
   & Ratio of correct detections\&recognitions to total number of predicted detections\&recognitions. Formula: $\frac{{\rm  TP}}{{\rm  TP} + {\rm  FP}}$ \cr \hline

   $\rm MOTA$
      & higher 
      & Multi-Object Tracking Accuracy~\cite{bernardin2008evaluating}. This measure combines three error sources: false positives, missed targets and identity switches. Formula: $1- \frac{{\rm  FN}+{\rm  FP}+{\rm  ID\,\,Sw.}}{{\rm  TP} + {\rm  FN}}$\cr \hline

   $\rm MOTP$
      & higher 
      &  Multiple Object Tracking Precision~\cite{bernardin2008evaluating} measures the average precision of target position predictions in multiple object tracking tasks.\cr \hline

    ${\rm Mostly\,\,  Tracked}$
   &higher &  Mostly tracked targets. The ratio of ground-truth trajectories that are covered by a track hypothesis for at least 80\% of their respective life span. \cr \hline
    ${\rm Partially\,\,  Matched}$
   & lower &  The percentage of ground-truth trajectories covered by a track hypothesis, within the range of 20\% to 80\% of their respective lifespan. \cr \hline
   
    ${\rm  Mostly\,\,  Lost}$
   & lower  &  Mostly lost targets. The ratio of ground-truth trajectories that are covered by a track hypothesis for at most 20\% of their respective life span. \cr \hline
   ${\rm  False\,\,  Postive(FP)}$
   & lower  &  The total number of false positives. \cr \hline
    ${\rm  False\,\,  Negative(FN)}$
   & lower  & The total number of false negative. \cr \hline
   ${\rm  True\,\, Postive(TP)}$
   & higher  &  The total number of true positives. \cr \hline
    ${\rm  True\,\, Negative(TN)}$
   & higher  & The total number of true negatives. \cr \hline
   ${\rm  ID\,\,Sw.}$
    & lower  & Number of Identity Switches~\cite{li2009learning}. The ID Switch metric measures the frequency of target ID changes in text tracking, representing the occurrences of target identity transitions within a sequence. \cr \hline
     \whline
\end{tabular}

%% file: table/detection.tex
\def\x{{$\scriptsize \times$}}
\scriptsize 
\arrayrulewidth0.5pt
\setlength{\tabcolsep}{1pt}
\begin{tabular}{c|c|c|ccc|c}
    \whline
    \multirow{2}{*}{Method} & \multirow{2}{*}{Backbone} & \multirow{2}{*}{Pretrain} &    \multicolumn{4}{c}{Video Text Detection/\%} \\
    \cline{4-7} 
    &  &  & Precision & Recall & F-measure & FPS
    \cr\shline \hline
	\hline

     EAST (size: 800)~\cite{wu2021bilingual} & VGG16 & COCO-Text  & 72.1 & 37.8 & 49.6 & 8.7 \\
     PSENet (size: 800)~\cite{psenet} & ResNet18 & COCO-Text  & 76.1 & 38.5 & 51.1 & 3.6 \\
     DB (size: 800)~\cite{liao2020real} & ResNet18 & COCO-Text  & 76.5 & 40.3 & 52.8 & 6.3 \\
     DB++ (size: 800)~\cite{liao2022real} & ResNet18 & COCO-Text  & 77.8 & 41.6 & 54.2 & 5.5 \\
     TransVTSpotter~(Q:200)~\cite{wu2021bilingual} & ResNet50 & COCO-Text  & 77.2 & 45.3 & 57.1 & 8.9 \\
     TransDETR~(Q:100)~\cite{wu2022end} & ResNet50 & - & 74.9  & 42.0 & 53.8 &  11.9 \\
     
     TransDETR~(Q:100)~\cite{wu2022end} & ResNet50 & COCO-Text  & 75.0 & 52.3 & 61.6 & 11.9 \\
     TransDETR~(Q:200)~\cite{wu2022end} & ResNet50 & COCO-Text  & 78.5 & 52.5 & 62.9 & 11.4 
     
    

    \cr\hline
     \whline
\end{tabular}
		

%% file: table/tracking.tex
\def\x{{$\scriptsize \times$}}
\scriptsize
\setlength{\tabcolsep}{2pt}
\begin{tabular}{c|c|c|c|ccc|ccc|c}
    \whline
	\multicolumn{2}{c|}{Method} & \multirow{2}{*}{$\rm Backbone$}   & \multirow{2}{*}{Pre.} & 
	\multicolumn{6}{c|}{Video Text Tracking/\%} & \multirow{2}{*}{FPS}  
	
    \cr\cline{1-2}\cline{5-10}  
    Detection &Tracking  &  & & ${\rm ID_{F1}}$ & $\rm MOTA$& $\rm MOTP$ & ${\rm M\mbox{-}T}$$\uparrow$ & ${\rm P\mbox{-}T}$$\uparrow$ & ${\rm M\mbox{-}L}$$\downarrow$ &  \cr\shline \hline
	\hline
	
     EAST&VMFT
     & VGG16 & COCO-Text & 32.1 & 24.8 & 76.1  & 2890 & 2670 & 7017  & 7.1 \\
     
   PSENet &VMFT
     & ResNet18 & COCO-Text & 33.8 & 24.5 & 77.3  & 3025 & 2509 & 7043 & 2.3

   \cr\hline
   \multicolumn{2}{c|}{YORO}
      & ResNet50 & COCO-Text & 33.2 & 23.4 & 73.5  & 985 & 1762  & 9893 & 5.4 \\
      
   \multicolumn{2}{c|}{TransVTSpotter~(Q:200) }
      & ResNet50 & COCO-Text & 50.1 & 31.0 & 77.2  & 2378 & 1892  & 8307 & 8.9 \\
   \multicolumn{2}{c|}{TransDETR~(Q:100) }
      & ResNet50 & - & 26.8 & 35.2 & 76.7  & 3191 & 2347  & 7039 & 11.9 \\
   \multicolumn{2}{c|}{TransDETR~(Q:200)}
      & ResNet50 & - & 47.1 & 30.4 & 77.1  & 2878 & 2091  & 7608 & 11.4 \\
   \multicolumn{2}{c|}{TransDETR~(Q:100) }
      & ResNet50 & COCO-Text & 52.7 & 36.0 & 77.8  & 3796 & 2484  & 6297 & 11.9 \\
    \multicolumn{2}{c|}{TransDETR~(Q:200) }
      & ResNet50 & COCO-Text & 55.2 & 39.3 & 77.4  & 4368 & 2505  & 5704 & 11.4 \\

     \whline
\end{tabular}

%% file: table/spotting.tex
\def\x{{$\scriptsize \times$}}
\scriptsize
\setlength{\tabcolsep}{2pt}
\begin{tabular}{c|c|c|c|c|ccc|ccc|c}
    \whline
	\multicolumn{3}{c|}{Method} & \multirow{2}{*}{$\rm Backbone$}   & \multirow{2}{*}{Pre.} & 
	\multicolumn{6}{c|}{Video Text Spotting/\%} & \multirow{2}{*}{FPS}  
	
    \cr\cline{1-3}\cline{6-11}  
    Detection &Tracking &Recognition &  & & ${\rm ID_{F1}}$ & $\rm MOTA$& $\rm MOTP$ & ${\rm M\mbox{-}T}$$\uparrow$ & ${\rm P\mbox{-}T}$$\uparrow$ & ${\rm M\mbox{-}L}$$\downarrow$ &  \cr\shline \hline
	\hline
	
     EAST & VMFT &CRNN
     & VGG16 & COCOText & 9.3 & -31.2 & 77.3  & 553 & 527  & 11497 & 6.7 \\
     
   PSENet & VMFT &CRNN
     & ResNet18 & COCOText & 10.2 & -31.0 & 77.3  & 536 & 587  & 11454 & 2.0

   \cr\hline
   \multicolumn{3}{c|}{TransDETR~(Q:100) }
      & ResNet50 & - & 9.8 & -16.4 & 77.9  & 407 & 404  & 11766 & 11.9 \\
   \multicolumn{3}{c|}{TransDETR~(Q:200) }
      & ResNet50 & - & 12.0 & -11.9 & 78.8  & 446 & 313  & 11818 & 11.4 \\
   \multicolumn{3}{c|}{TransDETR~(Q:100) }
      & ResNet50 & COCOText & 27.0 & -9.7 & 80.4  & 1458 & 780  & 10339 & 11.9 \\
    \multicolumn{3}{c|}{TransDETR~(Q:200) }
      & ResNet50 & COCOText & 25.5 & -1.3 & 81.3  & 1110 & 725  & 10742 & 11.4 \\

     \whline
\end{tabular}

%% file: table/generalizablity.tex
\def\x{{$\scriptsize \times$}}
\scriptsize
\setlength{\tabcolsep}{2pt}
\begin{tabular}{c|c|c|ccc|ccc}
    \whline
	 \multirow{2}{*}{Pre.} & \multirow{2}{*}{Train Set} & \multirow{2}{*}{Test Set} & 
	\multicolumn{6}{c}{Video Text Tracking/\%}   
	
    \cr\cline{4-9}  
    &  & & ${\rm ID_{F1}}$ & $\rm MOTA$& $\rm MOTP$ & ${\rm M\mbox{-}T}$$\uparrow$ & ${\rm P\mbox{-}T}$$\uparrow$ & ${\rm M\mbox{-}L}$$\downarrow$  \cr\shline \hline
	\hline
    COCO-Text & BOVText & DSText V2
       & 29.6 & 10.9 & 74.5  & 925 & 1801  & 9851  \\
    - & COCO-Text & DSText V2
       & 43.4 & 27.9 & 79.0 & 1533 & 1210  & 9834  \\   
    - & ICDAR2015 video & DSText V2
       & 44.8 & 29.1 & 73.1 & 1388 & 1329  & 9860  \\
       
    COCO-Text & ICDAR2015 video & DSText V2
       & 49.9 & 32.3 & 72.2  & 2406 & 1994  & 8177  \\
       
    COCO-Text & DSText V2 & DSText V2
       & 55.2 & 39.3 & 77.4  & 4368 & 2505  & 5704  

   \cr\hline
   - & DSText V2 & ICDAR 2015 video
       & 54.2 & 35.3 & 76.7  & 760 & 219  & 937  \\
    COCO-Text & DSText V2 & ICDAR 2015 video
       & 63.4 & 44.0 & 73.5  & 947 & 390  & 579  \\
     

     \whline
\end{tabular}

%% file: table/video_test_size.tex
\def\x{{$\scriptsize \times$}}
\scriptsize 
\arrayrulewidth0.5pt
\setlength{\tabcolsep}{2pt}
\begin{tabular}{c|c|c|c|c|c|c|c|c|c|c}
    \whline
    Video Name & 209 & 184 & 208 & 202 & 210& 122 & 228 & 199 & 214 & 156\\
     Text Area~(\# pixel) & 418.8
   &436.1 & 465.3 & 466.3 & 469.5& 506.9& 507.7& 518.0& 580.2& 600.6\\
     Text Number per Frame & 254.1 &  28.5 & 391.1 & 69.6 & 178.1 & 19.6 & 22.4 & 65.2 & 207.1 & 213.2\\
    ${\rm ID_{F1}}/\%$ & 21.8 & 42.4  & 31.9 & 40.9 & 47.1 & 58.4 & 43.7 & 58.2 & 37.0 & 70.0\\
    MOTA/\% & 21.4 & 24.6  & 20.8 & 23.6 & 33.3 & 48.6 & 22.3 & 21.1 & 29.1 & 51.5
    
    \cr \hline

	Video Name  & 128 & 239 & 162 & 121 & 135 & 175 & 192 & 125 & 143 & 216 \\
     Text Area~(\# pixel) &644.9 &666.4& 686.7& 688.4& 730.5& 735.2& 808.4& 812.7& 847.1& 872.7\\
     Text Number per Frame & 24.4 & 5.1 & 75.7 & 14.2 & 26.6 & 24.1 & 195.5 & 29.7 & 17.7 & 24.9\\
    ${\rm ID_{F1}}/\%$ & 67.7 & 58.6  & 63.6 & 55.0 & 50.0 & 57.1 & 24.0 & 66.7 & 51.7 & 54.8\\
    MOTA/\% & 54.0  & 38.9  & 48.3 &  32.4 & 39.4 & 36.2 & 27.1 & 40.7 &  24.3 & -9.6

    \cr \hline

	Video Name  & 145 & 149 & 126 & 174 & 112 & 232 & 222 & 211 & 133 & 197 \\
     Text Area~(\# pixel) & 900.9& 902.5& 987.9& 992.0& 998.8& 1017.5& 1018.7& 1033.4& 1039.5& 1068.3\\
     Text Number per Frame & 30.2 & 23.2 & 70.5 & 19.5 & 30.8 & 20.0 & 138.1 & 56.2 & 42.7 & 84.7\\
    ${\rm ID_{F1}}/\%$ & 68.9 &  51.7 & 54.5 & 65.3 &  80.6 & 50.5 & 45.0 & 51.5 & 47.6 & 62.8\\
    MOTA/\% & 48.6 & 46.8  & 32.4 & 46.9 & 68.4 & 29.8 & 41.5 & 39.4 & 31.1 & 45.2
    
    \cr \hline

	Video Name  & 196 & 171 & 167 & 223 & 237 & 172 & 132 & 231 & 220 & 218\\
     Text Area~(\# pixel) & 1128.8 & 1195.4 & 1196.8 & 1259.0 & 1312.1 & 1379.9 & 1385.7 & 1424.5 & 1511.8 & 1615.7 \\
     Text Number per Frame & 168.3 & 158.1 & 17.8 & 8.7 & 10.9 & 90.4 & 16.8 &32.2 & 60.4& 251.8\\
    ${\rm ID_{F1}}/\%$ & 32.5 & 27.7  & 62.2 &  66.9 & 66.3 & 49.2 & 59.7 & 37.6 & 59.4 & 42.0\\
    MOTA/\% & 42.7 & 27.9  & 36.8 & 42.0 & 51.6 & 33.3 & 29.0 & 17.2 & 46.2 & 50.2
    \cr \hline

	Video Name  & 230 & 242 & 136 & 105 & 108 & 144 & 178 & 134 & 114 & 123\\
     Text Area~(\# pixel) & 1769.3 & 1776.1 & 1783.3 & 2079.6 & 2366.3 & 2435.2 & 2959.7 & 3092.6 &  
    3310.9 & 6919.1\\
     Text Number per Frame & 6.3 & 11.8 & 24.6 & 25.8 & 24.1 & 14.5 & 79.8 & 12.5 & 18.3& 22.3\\
    ${\rm ID_{F1}}/\%$ & 62.4 &  63.4 & 58.6 & 63.9 & 68.8 & 64.6 & 51.9&  68.3 & 84.9 & 86.0\\
    MOTA/\% & 34.9 & 42.8  & 38.5 & 53.8 & 56.9 & 38.3 & 35.5 & 51.1 & 90.9 & 78.4
    \cr\hline
     \whline
\end{tabular}
		